\documentclass[sigconf,10pt]{acmart}

\usepackage{amsmath,amsfonts}
\usepackage{graphicx}
\usepackage{array}
\usepackage{booktabs}
\usepackage{tikz}
\usepackage{colortbl}
\usepackage{enumitem}
\usepackage[font=small]{caption}
\usepackage[noend]{algpseudocode}
\usepackage{algorithm}
\usepackage{natbib}
\usepackage{etoolbox}

\setcitestyle{numbers,sort&compress,square}

\newcommand\paraspace{\vspace*{0.5ex}}
\newcommand\parab[1]{\paraspace\noindent\textbf{#1}}

\newcommand{\figref}[1]{Fig.~\ref{#1}}
\newcommand{\appref}[1]{Section~\ref{#1}}
\newcommand{\apprefs}[2]{Sections~\ref{#1} and~\ref{#2}}

\input{config}

\usetikzlibrary{arrows.meta}

\begin{document}

\title{Diviner: Agentic RCA with Human-Governed Adaptation for Large-Scale Infrastructure}

\settopmatter{authorsperrow=4}

\author{Pradipta Ghosh}
\affiliation{\institution{Meta, Inc.}\country{USA}}
\email{iampradipta@meta.com}

\author{Gaurav Mitra}
\affiliation{\institution{Meta, Inc.}\country{USA}}
\email{gmitra@meta.com}

\author{Shripad Nadgowda}
\affiliation{\institution{Meta, Inc.}\country{USA}}
\email{shri.nadgowda@gmail.com}

\author{Akshil Shah}
\affiliation{\institution{Meta, Inc.}\country{USA}}
\email{akshilshah@meta.com}

\author{Junyu Liu}
\affiliation{\institution{Meta, Inc.}\country{USA}}
\email{jyliu@meta.com}

\author{Canturk Isci}
\affiliation{\institution{Meta, Inc.}\country{USA}}
\email{cisci@meta.com}

\author{Syamla Bandla}
\affiliation{\institution{Meta, Inc.}\country{USA}}
\email{shyamalab@gmail.com}

\author{Xinyao Hu}
\affiliation{\institution{Meta, Inc.}\country{USA}}
\email{xinyao.hu@meta.com}

\author{Onkar Deshpande}
\affiliation{\institution{Meta, Inc.}\country{USA}}
\email{odeshpande@meta.com}

\author{Anik Dasgupta}
\affiliation{\institution{Meta, Inc.}\country{USA}}
\email{anikdasgupta@meta.com}

\renewcommand{\shortauthors}{Ghosh et al.}

\begin{abstract}
Root-cause analysis (RCA) in large-scale distributed systems is challenging because a
single symptom can cross service, data, configuration, compute, and
organizational boundaries. Responders must assemble incomplete evidence and
coordinate experts who hold different parts of the causal path. Generic
language-model agents do not solve this problem through tool access alone: they
can use stale or mistimed context, interpret failed queries as absent symptoms,
confidently report unsupported state, and overload observability services
designed for human-paced use.

We present Diviner (\emph{Domain-Informed Verifiable Incident Navigation through
Evidence and Review}), an agentic RCA system operating at Meta and organized
around a four-stage reliability lifecycle.
\emph{Encode} uses a large corpus of reviewed incidents as a governed
operational curriculum, combining recurring failure patterns with service-owner
expertise to produce scoped, versioned skills, workflows, typed tools, and an
incident knowledge graph. \emph{Execute} uses an adaptive Lead and
context-isolated workers to apply relevant artifacts to point-in-time evidence
and maintain evidence-linked hypotheses. \emph{Evaluate} compares saved
investigation trajectories with responder-reviewed outcomes to measure exact
cause, useful direction, and Groundedness. \emph{Adapt} is a human-gated process
that uses recurring gaps and on-call feedback to derive focused
operational-knowledge updates, admitting them only after replay, quality checks,
and owner approval.

In observational cohort comparisons within Meta's large-scale recommendation
and ranking infrastructure, Diviner-assisted incidents were associated with
approximately 52\% lower detection-to-mitigation time and approximately 15\%
lower pre-mitigation operational effort. Across scored investigations, Diviner identifies the exact
root cause in 45\% of cases, provides a useful causal direction in 60\%, and
achieves 80\% mean Groundedness.

\footnote{\textbf{Generative-AI disclosure.} The authors used Codex, Claude, and Gemini
for literature discovery, language editing, and initial figure layouts. The
authors verified all citations, claims, results, and figures.}

\end{abstract}

\maketitle
\vspace{-0.1in}

\section{Introduction}
\label{sec:introduction}

\begin{figure*}[!t]
\centering
\includegraphics[width=\textwidth]{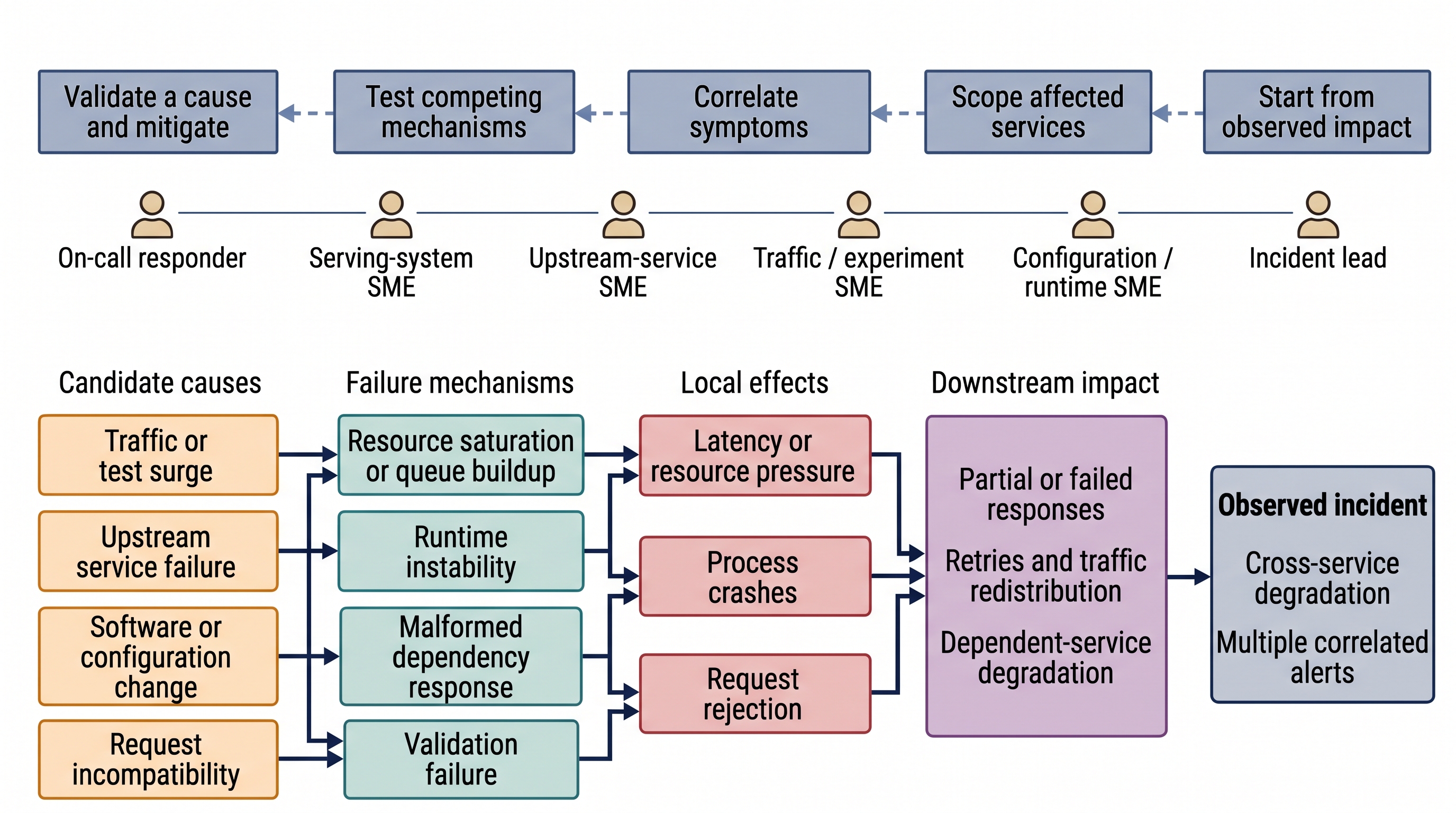}
\caption{Anatomy of a cross-service incident investigation. Solid arrows show
representative paths from candidate causes through failure mechanisms and local
effects to downstream impact. The dashed path above shows responders tracing
the incident in the reverse direction, while service experts examine different
parts of the system in parallel.}
\label{fig:incident_anatomy}
\end{figure*}

In large-scale distributed systems, the component exhibiting a failure is often
far removed from the component that caused it. A request may traverse data
pipelines, services, experimentation, configuration, shared
compute, networking, and storage before a user-visible symptom appears. These
components evolve independently, so a fault in one layer may surface later as a
statistical shift or failure in another~\cite{sillito2020failures,beyer2016sre}.
Root-cause analysis (RCA) requires responders to form competing hypotheses,
find discriminating evidence, and redirect the investigation as observations
arrive. Even before the exact cause is known, localizing the right subsystem,
refuting an alternative, or identifying the next test can accelerate
mitigation. We call this evidence-backed progress \emph{useful direction}.
Figure~\ref{fig:incident_anatomy} illustrates both how a fault propagates across
service boundaries and how responders trace the resulting impact back toward
candidate causes.

Large language model (LLM) agents create an opportunity to automate parts of
this process.
They can read procedures, call diagnostic tools, pursue several hypotheses, and
synthesize evidence. RCA, however, is not an ordinary task-completion problem.
Unlike a coding agent, an incident investigator does not receive a stable
repository, a deterministic test suite, or a clean corpus of tasks paired with
known answers. The necessary evidence changes while the incident unfolds, and
several investigation paths may produce useful progress without reaching the
same final explanation. Recent evaluations of RCA agents, IT operations, and
on-call tasks show that strong benchmark or coding performance does not
necessarily translate into reliable incident investigation
\cite{xu2025openrca,chen2025aiopslab,itbench2025,gong2026oncall}.
The necessary expertise is distributed across service owners, and a plausible
unsupported answer can redirect scarce responder attention. Reliable
assistance therefore requires more than a capable model or a large tool
catalog: it requires \emph{governed knowledge, explicit evidence semantics,
bounded execution, and human-governed updates from reviewed outcomes}.

We present Diviner (\emph{Domain-Informed Verifiable Incident Navigation through
Evidence and Review}), an agentic RCA system operating at Meta in large-scale
distributed infrastructure. Diviner treats each incident as one
turn of a continuing operational process rather than an isolated agent run.
We organize this process as a four-stage reliability lifecycle. \emph{Encode}
turns recurring corpus patterns and expert knowledge into governed artifacts;
\emph{Execute} applies them in adaptive, evidence-producing investigations
using a custom multi-agent harness;
\emph{Evaluate} compares saved trajectories with reviewed outcomes to measure
investigation quality; and \emph{Adapt} uses
reviewed gaps to derive human-gated updates for future operations.
Our thesis is that reliable RCA agents require governed uncertainty reduction:
reasoning may remain adaptive, but permitted knowledge, evidence admission, and
durable knowledge changes must remain scoped and attributable.

This lifecycle structures Diviner's design. The paper makes three contributions:
\begin{enumerate}[leftmargin=1.4em,itemsep=2pt]
    \item \textbf{Governed operational knowledge.} Diviner combines a large
    reviewed incident corpus with service-owner expertise to produce scoped,
    versioned failure signatures, workflows, skills, typed tools, and
    knowledge-graph relationships. Applicability, provenance, evaluation, and
    owner approval govern what later investigations may use
    (Sections~\ref{sec:operational_curriculum} and~\ref{sec:encoding}).

    \item \textbf{A four-stage RCA-focused agent lifecycle.} Diviner connects
    Encode, Execute, Evaluate, and Adapt through bounded orchestration, isolated
    workers, evidence-linked hypotheses, trajectory evaluation, and
    human-gated knowledge updates
    (Sections~\ref{sec:system_design} and~\ref{sec:implementation}).

    \item \textbf{Operational evaluation and experience.} We report
    investigation quality and latency, incident outcomes, controlled
    configuration comparisons, early knowledge-update measurements, and lessons
    from sustained operational use
    (Sections~\ref{sec:evals} and~\ref{sec:lessons}).
\end{enumerate}

At Meta, Diviner has supported several thousand incidents, including several hundred
high-severity incidents, and runs several thousand alert-triggered
investigations each month. It identified the exact root cause in 45\% of scored
cases, provided a useful causal direction in 60\%, and achieved 80\% mean
Groundedness. Across onboarded cohorts, assisted incidents were associated with
approximately 52\% lower detection-to-mitigation time and approximately 15\%
lower pre-mitigation operational effort. Diviner produced a first useful,
medium-confidence hypothesis after 7.0 minutes at p50 and 13.7 minutes at p95;
completed investigations reached a final conclusion after approximately
15 minutes at p50 and 24.0 minutes at p95.

\section{Anatomy of a Cross-Service Incident}
\label{sec:incident_anatomy}

This section synthesizes a class of real incidents involving the same
model-serving symptom: sudden increases in rejected requests and recurrent
process terminations. Although the investigations followed a common structure,
the affected services and root causes varied. Figure~\ref{fig:incident_anatomy}
summarizes the recurring failure paths and the reverse investigation followed
by responders.

\subsection{One Symptom, Several Plausible Causes}

Rejected requests and process terminations could result from upstream-service
failures, test or traffic activity, software or configuration changes, or
incompatible requests (Figure~\ref{fig:incident_anatomy}). Each possibility
required different evidence and mitigation, so the symptom alone could not
identify the cause.

To debug these incidents, the initial responder established the affected
request paths, services, regions, and time windows. A serving-system expert
aligned the symptom with recent binaries, configurations, resource behavior,
and request characteristics. As evidence crossed a boundary, the relevant
upstream, traffic, experimentation, or configuration experts joined. The
investigation accumulated partial results, competing explanations, and evidence
links, consistent with prior studies of incident response
\cite{sillito2020failures,beyer2016sre}.

\subsection{Investigation as Distributed Causal Inference}

Responders did not follow a fixed checklist to completion. They formed
competing hypotheses and selected observations that could distinguish among
them. A time-aligned software change strengthened one branch but did not prove
it; a rollback that failed to restore service weakened the claim that the
change was the only active mechanism. A query returning no matching records
could be counter-evidence only when the target, time range, permissions, and
result status were valid. A timeout or partial response established no such
fact.

Responders therefore had to reconstruct a causal timeline across interdependent
systems, coordinate experts across the request path, and preserve which
hypotheses were supported, refuted, or unresolved. Mitigation remained a human
decision, while evidence gathered during recovery continued to refine the
diagnosis.

\subsection{What Repeats and What Changes}

Request paths, common upstream branches, useful comparisons, and discriminating
checks carried across recurrences. The affected service, recent changes,
traffic source, and request form did not. A later incident therefore required
neither starting from scratch nor assuming that the previous cause had
returned. What carried forward was the prior investigation path and its tests,
not its conclusion.

This distinction motivates an RCA agent. Much of the evidence collection and
hypothesis testing is systematic and can begin before every expert assembles,
but the investigation path must remain adaptive and grounded in current state.
The next section generalizes this observation using the incident corpus and
derives the requirements for a reliable RCA agent.

\section{The Case for RCA Agents}
\label{sec:background}
\label{sec:case_for_rca_agents}

Section~\ref{sec:incident_anatomy} illustrates one class of cross-service
investigation. This section uses a broader incident corpus to show why RCA
requires adaptive agents, why generic agents are insufficient, and which
requirements follow.

\subsection{Operational Diversity Defeats Fixed Procedures}

\begin{figure}[t]
\centering
\includegraphics[width=\columnwidth]{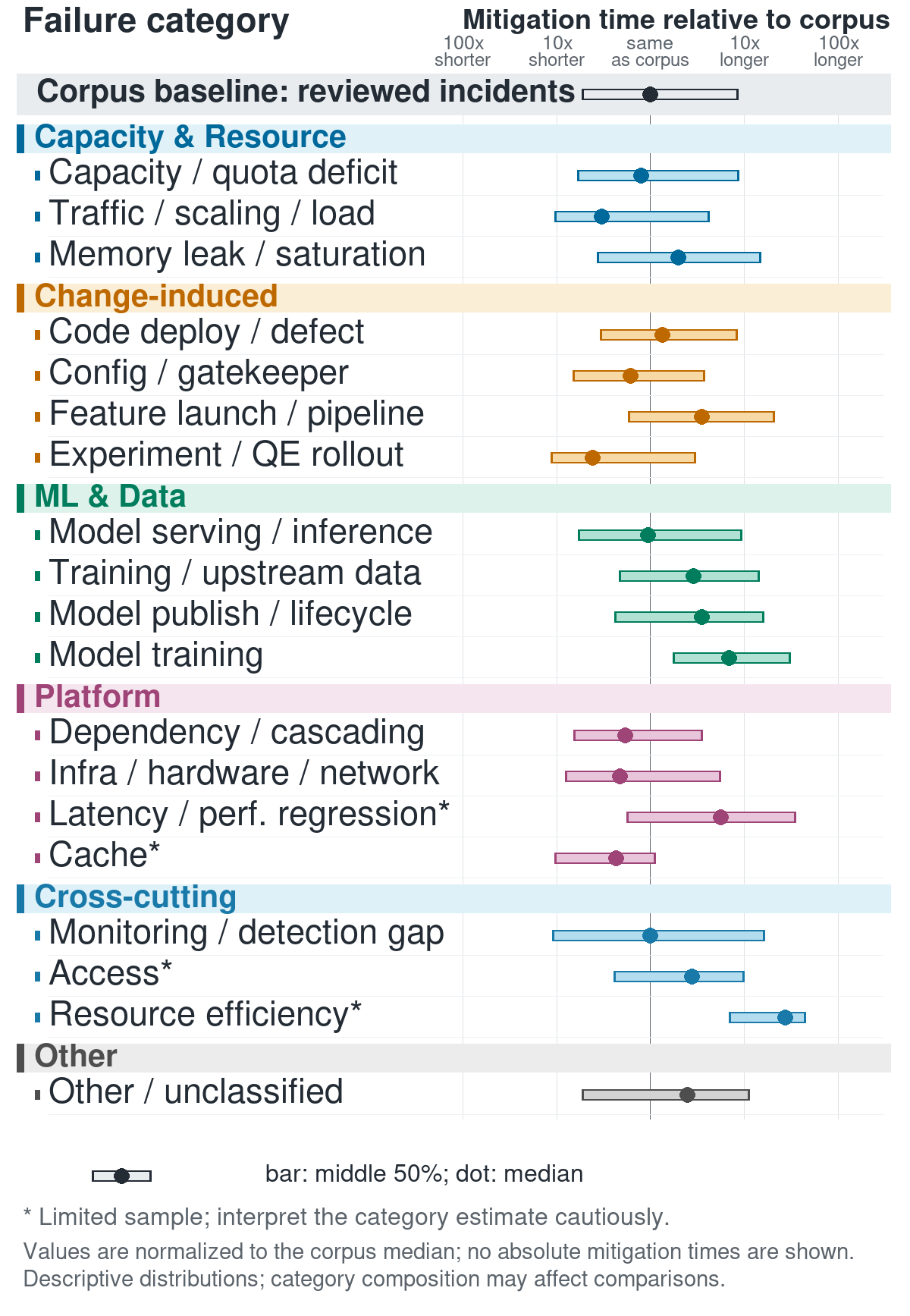}
\caption{Relative mitigation time by cause in reviewed
ranking-infrastructure incidents. Values are normalized to the corpus-wide
median; bars span Q1--Q3 and dots mark category medians on a logarithmic axis.}
\label{fig:incident_categories}
\end{figure}

We analyzed several thousand reviewed incidents from recommendation and ranking
infrastructure. Figure~\ref{fig:incident_categories} shows that median mitigation time
varies by more than an order of magnitude across and within cause categories.
Some incidents are mitigated by reverting a change or shifting traffic. Others
require data repair, model retraining or republishing, and validation across
multiple services. Mitigation time includes diagnosis, coordination, and these
recovery steps.

The distribution motivates two coupled objectives: \textit{reduce time from detection
to mitigation (TDM) and reduce pre-mitigation operational effort.}
The latter measures estimated responder effort accumulated before mitigation.
Table~\ref{tab:terminology} in \appref{app:terminology} defines the
operational and evaluation measures used throughout the paper.

The variation also rules out a single diagnostic path. A reliable RCA agent must
establish scope, maintain competing explanations, select checks that distinguish
among them, and revise the investigation when evidence contradicts its current
plan. When the exact cause remains unresolved, it should still return a useful
causal direction or the next discriminating test.

\subsection{Educating the Agent}
\label{sec:operational_curriculum}

Reviewed incidents reveal recurring failure patterns and the evidence that
distinguishes similar symptoms with different causes. Automated processing
structures these records into failure signatures and knowledge-graph
relationships. Service owners provide the complementary procedural knowledge by
encoding diagnostic checks as reusable skills, workflows, and typed tools.

This knowledge cannot be captured once and treated as complete. Topology,
ownership, telemetry semantics, and diagnostic practice differ by service and
change over time. A central platform team cannot maintain every procedure, so
authoring must remain distributed among service owners. The resulting artifacts
must remain owned, scoped, validated, versioned, measured, and retired when they
become stale.

We call this process \emph{educating the agent}: operational knowledge is updated
through external artifacts rather than model retraining. Prior incidents provide
hypotheses and diagnostic tests, but they do not establish the current system
state. An RCA agent must verify each hypothesis using current evidence.

\subsection{Conventional Triage Mechanisms}
\label{app:conventional_automation}

Conventional automation accelerates individual stages of an investigation, but
each class assumes that part of the diagnostic path is known in advance.
Operational studies identify configuration, operator action, and poorly tested
recovery paths as recurring contributors
\cite{oppenheimer2003internet,yuan2014simple}.

\parab{Monitoring and runbooks.} Alerts detect known symptoms, dashboards align
common signals, and runbooks automate repeatable checks. They work well when a
team has anticipated the failure mode and encoded the right identifiers. They
cannot choose a new cross-system query when evidence invalidates the expected
path.

\parab{Cross-signal diagnosis.} Tracing reconstructs request paths, while
statistical and learned methods rank anomalous components. Mystery Machine,
Eadro, and DiagFusion show the value of combining traces, logs, metrics, and
dependency structure
\cite{chow2014mystery,lee2023eadro,zhang2023diagfusion}. Their analyses depend on
defined telemetry schemas and failure classes; gray failures can still make
observers disagree about component health, while request fan-out amplifies rare
local delays \cite{huang2017gray,dean2013tail}.

\parab{Experience retrieval.} Systems such as ESRO combine live alerts with
graphs of prior outage reports to retrieve likely causes and mitigations
\cite{chakraborty2023esro}. This history can narrow the search, but a similar
incident is a hypothesis source, not evidence about the current system.

Together, these mechanisms detect symptoms, automate known checks, combine
selected signals, and retrieve past experience. They do not coordinate the full
open-ended investigation. Responders must still choose among heterogeneous
tools, redirect the plan, reconcile partial results, and preserve why they
reached a conclusion \cite{sillito2020failures,beyer2016sre}.

\subsection{Why RCA Requires Adaptive Agents}
\label{sec:background_agents}

An RCA agent can go beyond executing such a fixed path. It can read relevant
procedures, choose among heterogeneous tools, pursue independent hypotheses,
reconcile results across systems, and revise its next step as evidence changes.
This adaptability matters because neither the cause nor the useful investigation
path is known when an incident begins.

\parab{RCA is not a coding task.} Coding agents commonly work in a stable
workspace with a repeatable build or test signal. Operational RCA starts from
an incomplete symptom in a changing environment and has neither a deterministic
path nor verifier~\cite{gong2026oncall}. Reliable investigators may follow different evidence paths,
while a lucky final answer can conceal an unreliable trajectory.

\subsection{Why Generic Agents Are Not Enough}

Adaptive investigation creates an asymmetric risk: an unresolved result leaves
the responder investigating, whereas plausible but unsupported guidance can
divert attention and delay mitigation. Prompting the model to be cautious does
not solve this problem. In common model-training and benchmark evaluations,
accuracy-based objectives can reward guessing rather than abstaining when
evidence is insufficient~\cite{kalai2026accuracy}. Larger models can produce
plausible false claims~\cite{lin2022truthfulqa}, and self-correction without new
external evidence can reduce accuracy~\cite{huang2024selfcorrect}.

Operational tools amplify the risk because telemetry and change systems differ
in clocks, permissions, retention, and failure semantics. A timeout, denial,
partial response, or wrong time range is not evidence that a symptom is absent.
RCA therefore needs explicit failure and abstention states plus deterministic
controls over evidence admission, permissions, concurrency, and resource use.

A large incident corpus also does not automatically provide a reliable
evaluation set. Postmortems often combine evidence available during response
with conclusions learned later; telemetry may expire, tools and permissions may
change, and some incidents never receive a definitive root-cause label.
Evaluation must therefore reconstruct the information available at incident
time and assess whether the agent found the exact cause, narrowed the
investigation usefully, and supported its claims with valid evidence.

Prior systems provide incident context, executable guides, tool-augmented
planning, evidence-grounded hypotheses, and evolving operational skills
\cite{chen2024rcacopilot,an2024nissist,wang2024rcagent,jiang2024xpert,
wei2026agentrca,liu2026skillforge,huang2026opsharness}. These capabilities
advance individual investigations but do not jointly govern knowledge
applicability, evidence-based hypothesis updates, trajectory evaluation, and
the reuse of reviewed outcomes.

\subsection{Requirements for Reliable RCA Agents}

The responsibilities and risks above yield the six requirements in
Table~\ref{tab:requirements}.

\begin{table*}[t]
\centering
\caption{Requirements for AI-assisted operational investigation.}
\label{tab:requirements}
\small
\resizebox{\textwidth}{!}{
\begin{tabular}{@{}p{2.0cm}p{6.4cm}p{7.7cm}@{}}
\toprule
\textbf{Requirement} & \textbf{Why it is needed} & \textbf{Design obligation} \\
\midrule
Grounding & Plausible but absent, stale, or unsupported details can misdirect responders. & Link findings to current, attributed evidence; distinguish empty data from tool failure; abstain when support is insufficient. \\
\midrule
Control and load & Adaptive planning and machine-speed queries can amplify a wrong branch, retry loop, or backend load. & Keep the plan revisable while bounding capabilities, concurrency, retries, context, deadlines, and resource use. \\
\midrule
Knowledge & Models lack local state, while procedures, ownership, and interfaces drift. & Encode owned, versioned artifacts and track their applicability and quality. \\
\midrule
Evaluation & A fluent answer can hide a lucky guess, unsupported reasoning, or wasted work. & Evaluate outcome and trajectory, including direction, grounding, failed checks, and efficiency. \\
\midrule
Performance & Assistance must arrive before responders find the same evidence. & Measure time to supported findings and final conclusions, including tail latency. \\
\midrule
Action boundary & Tools access sensitive systems, but the model has no authority to change them. & Preserve service authorization and auditability; leave mitigation to responders. \\
\bottomrule
\end{tabular}
}
\end{table*}

Together, these requirements define the objective: reach an actionable,
evidence-supported causal hypothesis sooner without weakening provenance,
reliability, or human authority. They call for an RCA-specific lifecycle rather
than relying on prompting and unrestricted tool access.

\subsection{Closing the Loop}

For the incident class in Section~\ref{sec:incident_anatomy}, experts encoded
checks for upstream behavior, traffic, changes, and request compatibility.
Later investigations applied them to current evidence, while post-mitigation
review refined missing and service-specific branches. After several cycles,
Diviner isolated the upstream source within minutes in subsequent incidents and
returned evidence-linked findings for responder validation. This loop makes
expertise reusable without treating precedent as proof; Section
\ref{sec:system_design} presents its design.

\section{System Design: The Four-Stage Lifecycle}
\label{sec:system_design}
\label{sec:system_overview}
The incident anatomy in Section~\ref{sec:incident_anatomy} exposed a
cross-service search problem, and Section~\ref{sec:background} showed why
neither a static runbook nor a generic tool-using agent is sufficient.
Table~\ref{tab:requirements} translates those risks into concrete constraints
on Diviner's knowledge, execution, evaluation, performance, and responder
interfaces.

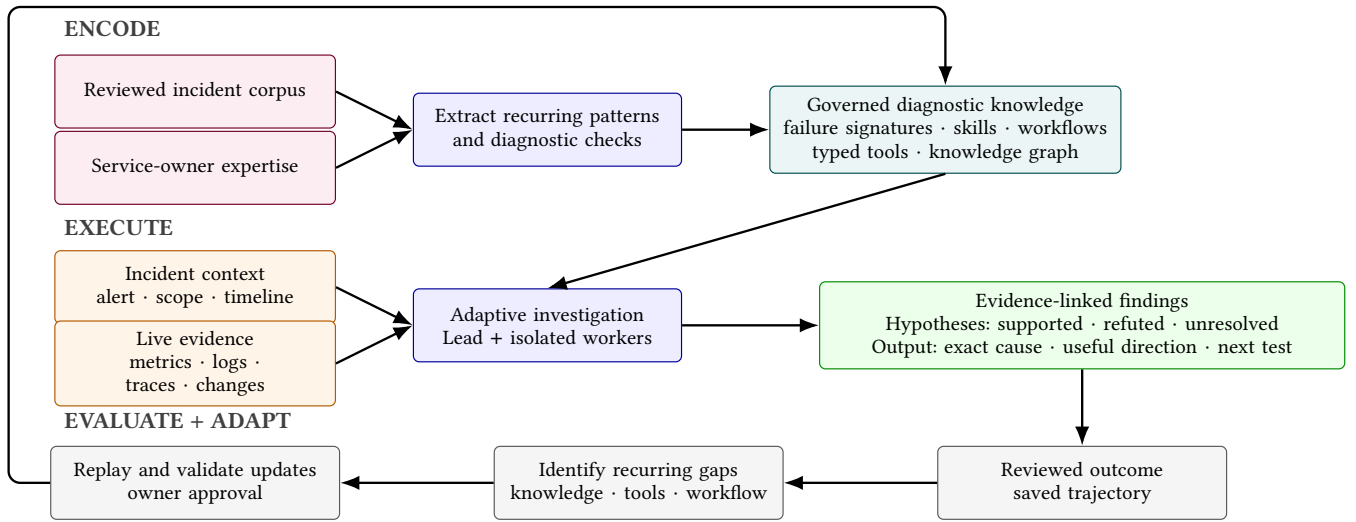
\begin{figure*}[t]
\centering
\resizebox{\textwidth}{!}{\begin{tikzpicture}[
    x=1cm,
    y=1cm,
    >=Latex,
    source/.style={draw=purple!60!black, rounded corners=2pt, fill=purple!7,
                   align=center, text width=3.15cm, minimum height=0.88cm,
                   font=\scriptsize},
    transform/.style={draw=blue!60!black, rounded corners=2pt, fill=blue!7,
                      align=center, text width=3.0cm, minimum height=0.88cm,
                      font=\scriptsize},
    knowledge/.style={draw=teal!60!black, rounded corners=2pt, fill=teal!8,
                      align=center, text width=4.0cm, minimum height=1.05cm,
                      font=\scriptsize},
    live/.style={draw=orange!70!black, rounded corners=2pt, fill=orange!9,
                 align=center, text width=3.15cm, minimum height=0.88cm,
                 font=\scriptsize},
    output/.style={draw=green!55!black, rounded corners=2pt, fill=green!8,
                   align=center, text width=3.55cm, minimum height=0.88cm,
                   font=\scriptsize},
    review/.style={draw=gray!70!black, rounded corners=2pt, fill=gray!8,
                   align=center, text width=3.25cm, minimum height=0.88cm,
                   font=\scriptsize},
    band/.style={font=\bfseries\footnotesize, anchor=west, text=black!75}
]

% Offline knowledge construction.
\node[band] at (-1.7,2.68) {ENCODE};
\node[source] (corpus) at (0,1.92)
    {Reviewed incident corpus};
\node[source] (experts) at (0,1.00)
    {Service-owner expertise};
\node[transform] (patterns) at (4.25,1.46)
    {Extract recurring patterns\\and diagnostic checks};
\node[knowledge] (artifacts) at (9.05,1.46)
    {Governed diagnostic knowledge\\failure signatures $\cdot$ skills $\cdot$ workflows\\typed tools $\cdot$ knowledge graph};
\draw[->,thick] (corpus.east) -- (patterns.west);
\draw[->,thick] (experts.east) -- (patterns.west);
\draw[->,thick] (patterns.east) -- (artifacts.west);

% Online investigation.
\node[band] at (-1.7,0.28) {EXECUTE};
\node[live] (alert) at (0,-0.44)
    {Incident context\\alert $\cdot$ scope $\cdot$ timeline};
\node[live] (state) at (0,-1.36)
    {Live evidence\\metrics $\cdot$ logs $\cdot$ traces $\cdot$ changes};
\node[transform] (investigate) at (4.25,-0.90)
    {Adaptive investigation\\Lead $+$ isolated workers};
\node[output,text width=6.1cm,minimum height=1.05cm] (finding) at (10.70,-0.90)
    {Evidence-linked findings\\Hypotheses: supported $\cdot$ refuted $\cdot$ unresolved\\Output: exact cause $\cdot$ useful direction $\cdot$ next test};
\draw[->,thick] (alert.east) -- (investigate.west);
\draw[->,thick] (state.east) -- (investigate.west);
\draw[->,thick] (artifacts.south) -- (investigate.north);
\draw[->,thick] (investigate.east) -- (finding.west);

% Post-incident evaluation and update.
\node[band] at (-1.7,-2.05) {EVALUATE $+$ ADAPT};
\node[review] (outcome) at (10.70,-2.80)
    {Reviewed outcome\\saved trajectory};
\node[review] (gaps) at (5.35,-2.80)
    {Identify recurring gaps\\knowledge $\cdot$ tools $\cdot$ workflow};
\node[review] (update) at (0,-2.80)
    {Replay and validate updates\\owner approval};
\draw[->,thick] (finding.south) -- (outcome.north);
\draw[->,thick] (outcome.west) -- (gaps.east);
\draw[->,thick] (gaps.west) -- (update.east);
\draw[->,thick,rounded corners=4pt] (update.west) -- (-2.25,-2.80)
    -- (-2.25,2.94) -- (9.05,2.94) -- (artifacts.north);

\end{tikzpicture}}
\caption{Diviner's four-stage lifecycle. Encode builds governed diagnostic
knowledge; Execute combines it with live evidence; Evaluate identifies gaps in
the saved trajectory; and Adapt publishes replayed, owner-approved updates.}
\label{fig:corpus_system_overview}
\end{figure*}

Diviner follows one four-stage lifecycle, shown in
Figure~\ref{fig:corpus_system_overview}. \emph{Encode} turns reviewed incidents
and service-owner expertise into governed diagnostic knowledge. \emph{Execute}
applies that knowledge to current evidence through a Lead and focused workers.
After mitigation, \emph{Evaluate} compares the saved trajectory with the
reviewed outcome, and \emph{Adapt} converts recurring gaps into replayed,
owner-approved artifact updates.

Across the lifecycle, claims must cite retrievable evidence, agents may use only
registered knowledge and tools, and humans retain authority over mitigation and
shared knowledge.

\subsection{Encode: From Incident Corpus to Governed Knowledge}
\label{sec:encoding}

Encode turns corpus patterns and service-owner expertise into governed,
versioned diagnostic knowledge. Automated processing distills each reviewed
incident into a failure signature (a compact summary of its context, symptoms,
cause, distinguishing evidence, and mitigation) and adds it to an Incident
Knowledge Graph (IKG). Service owners contribute procedural knowledge as skills,
workflows, and typed tools. Registries record the scope, owner, version,
provenance, validation, and enablement of these artifacts so the runtime can
retrieve only those applicable to the current incident.

\subsubsection{Skills and Workflows}
\label{sec:encoding_standards}
\label{sec:encoding_skills}
\label{sec:encoding_registry}
\label{sec:encoding_workflows}

A \emph{skill} is the smallest reusable diagnostic unit: it specifies how to
perform one check, interpret its observations, and handle expected failure
conditions. Diviner represents skills using the Agent Skills
format~\cite{agentskills2026spec}, with instructions, metadata, and optional
scripts.

A \emph{workflow} is the full investigation strategy for an incident class. It
connects skills in a directed acyclic graph. Nodes reference skills or decision
points; edges encode prerequisites, parallel branches, evidence-dependent next
steps, and convergence. This structure captures where experts begin, which
competing causes they test, and which observations justify continuing or pruning
a branch (see \figref{fig:incident_anatomy}). Skills remain independently
reusable, while workflows provide the cross-service coordination that a
collection of isolated procedures lacks.

The registries record the artifact versions selected for each investigation,
supporting later evaluation and rollback. Section~\ref{sec:execution} explains
how the Lead applies and adapts a workflow; Section~\ref{sec:impl_registry} and
\appref{app:skill_registry} give registry details.

\subsubsection{Tools for Evidence Collection}
\label{sec:mcp}
\label{sec:encoding_tools}

Diviner packages recurring analyses as purpose-built Model Context Protocol
(MCP) tools~\cite{mcp2025spec}. Each declares ownership, scope, permissions,
output bounds, and result types; the runtime records its version, parameters,
target, time range, and status. Section~\ref{sec:impl_tools} and
\appref{app:tool_plane} give details.

\subsubsection{Historical and Live Context}
\label{sec:encoding_incident_memory}
\label{sec:encoding_kg}

The IKG supplies versioned historical precedents, while separate retrieval paths
supply current dependencies, ownership, changes, configurations, experiments,
and ongoing incidents. Historical matches can suggest a hypothesis or test, but
only current evidence can support a finding. The following subsection describes
how the IKG is constructed and queried.

\subsubsection{IKG Construction and Retrieval}
\label{app:ikg}

\begin{figure}[t]
\centering
\includegraphics[width=\linewidth]{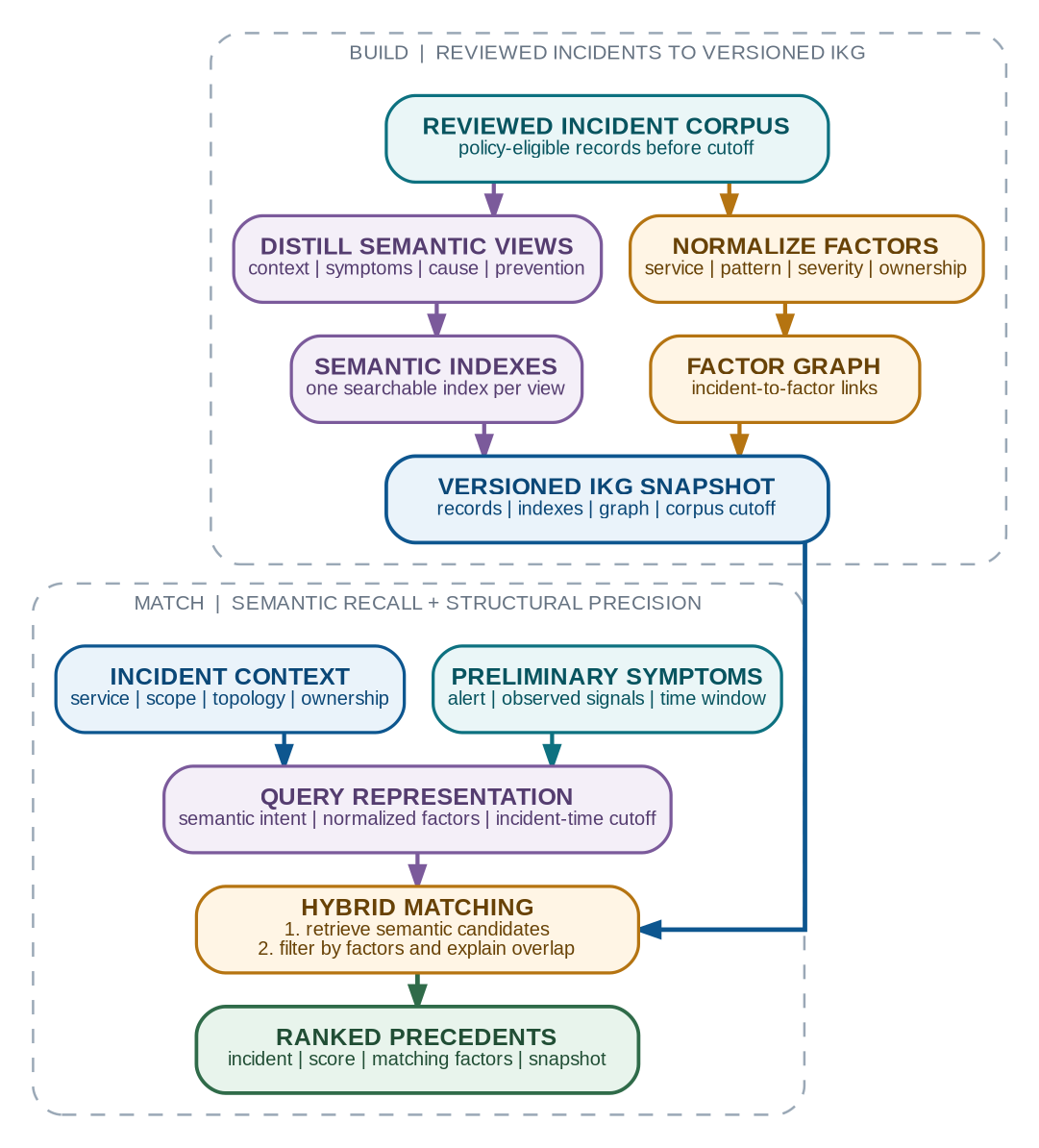}
\caption{IKG construction and matching. Reviewed incidents are distilled into
semantic views and normalized factors, which form a versioned dual index. At
runtime, preliminary symptoms and incident context form a query; semantic
retrieval supplies candidates and shared factors constrain and explain the
matches. Returned precedents remain hypotheses until verified with live tools.}
\label{fig:ikg_pipeline}
\end{figure}

The Incident Knowledge Graph (IKG) converts reviewed incident history into
retrievable diagnostic precedents. It is not a live service-topology model and
does not establish the current root cause. Instead, it represents what was
observed in a prior incident, how responders distinguished competing causes,
and which resolution was ultimately reviewed. Figure~\ref{fig:ikg_pipeline}
summarizes its construction, serving, and verification boundaries.

\parab{Construction.} An eligible incident is first distilled into a compact
failure signature rather than indexed as one undifferentiated postmortem. The
pipeline produces separate semantic views for the operational context, observed
symptoms, reviewed causal mechanism, and mitigation or prevention guidance.
The index encodes each view independently with DRAMA-1B as a
2,048-dimensional dense vector. The vectors are stored in four FAISS indexes,
one for each semantic view, together with the incident identifier, view type,
snapshot version, and corpus cutoff. Separating the indexes allows a symptom
query to retrieve incidents with similar observations without requiring their
final causes to be textually similar, and prevents a long incident narrative
from dominating the match.

In parallel, the pipeline normalizes structured factors such as service scope,
failure family, severity class, region, and ownership. The graph links each
incident record to these factors; it does not infer a live dependency graph.
The incident record, semantic indexes, factor links, and corpus cutoff are
published together as a versioned IKG snapshot. An investigation remains pinned
to its selected snapshot so later corpus additions or corrections do not
silently change its context.

\parab{Retrieval and serving.} At runtime, the retrieval interface combines the
alert, preliminary symptoms, affected scope, and time window into semantic
queries and normalized factors. DRAMA-1B encodes each semantic query in the same
vector space, and FAISS nearest-neighbor search retrieves candidates from the
corresponding indexes. The candidate lists are merged by incident identifier;
factor overlap then filters or reranks them and explains the match through
shared service, failure-family, or ownership attributes. Only a small ranked
set is hydrated for the agent, together with the relevant failure signature,
matched semantic view, matching factors, retrieval score, snapshot version,
and cutoff. This keeps the full corpus out of the prompt and makes the source
of each precedent inspectable.

Corpus admission and result hydration apply separate access checks. Historical
evaluation uses a snapshot preceding the target incident and excludes later
revisions of that incident. Because semantic similarity and factor overlap can
both be misleading, a returned precedent is only a hypothesis generator. The
Lead must verify it against current metrics, changes, configuration, request
properties, or dependency state before using it in a conclusion.

\subsection{Execute: Evidence-Grounded Investigation}
\label{sec:execution}

Diviner implements an event-driven Lead--worker architecture on
LangChain~\cite{langchain2022}. When an incident triggers Diviner, the Lead
first establishes the affected scope and time window. The runtime uses this context and the
preliminary symptoms to search registry metadata for an applicable workflow and
to retrieve relevant IKG precedents. The workflow and precedents seed the
investigation, but only evidence from the current incident can support a
finding.

Tool results, worker updates, and responder input return control to the Lead,
allowing it to revise the plan as the investigation progresses.

\subsubsection{Adapt the Plan as Evidence Arrives}

The Lead begins by instantiating the registered workflow as the investigation
plan. Each step states a question, target, time window, expected observation,
and diagnostic skill. Independent steps run concurrently in isolated workers,
each receiving only its question, relevant context, skill, and required tools.
Workers return structured observations rather than sharing complete
transcripts. As evidence arrives, the Lead may add new branches, reprioritize
existing ones, or prune branches contradicted by the evidence. These adaptations
apply only to the current investigation and do not rewrite the registered
workflow. The runtime records the workflow and skill versions used by each plan
step. Only post-incident review can turn a successful live-plan change into a
new workflow version for later investigations.

\subsubsection{Record Evidence and Uncertainty}
\label{sec:execution_grounding}

Diviner records an explicit set of hypotheses rather than relying on the chat
history as memory. Each hypothesis is marked as supported, refuted, or
unresolved and links to the observations used to assess it. Every observation
records its source, scope, time range, tool outcome, and the skill that requested
it.

Tool outcomes distinguish data, valid empty results, partial results, and
failures. This distinction prevents a failed query from being interpreted as an
absent symptom. Only valid observations from the intended target and time range
can support or refute a hypothesis. Conflicting evidence and unresolved tests
remain visible in the final state.

\subsubsection{Review the Result and Bound the Search}
\label{sec:skeleton}

When the Lead proposes a conclusion, the runtime Review Agent checks its cited
evidence, conflicts, and unresolved tests. It retains claims supported by the
record and returns unsupported claims as concrete questions for another focused
check. A limited number of review rounds prevents this process from continuing
indefinitely.

If evidence remains incomplete, Diviner returns a partial assessment that may
identify a useful direction, refute a branch, or recommend the next test.
Deadlines, concurrency limits, cancellation, permissions, and checkpoints bound
the investigation. Diviner remains advisory, and responders choose whether and
how to mitigate the incident.

\begin{figure}[t]
\centering
\includegraphics[width=\columnwidth]{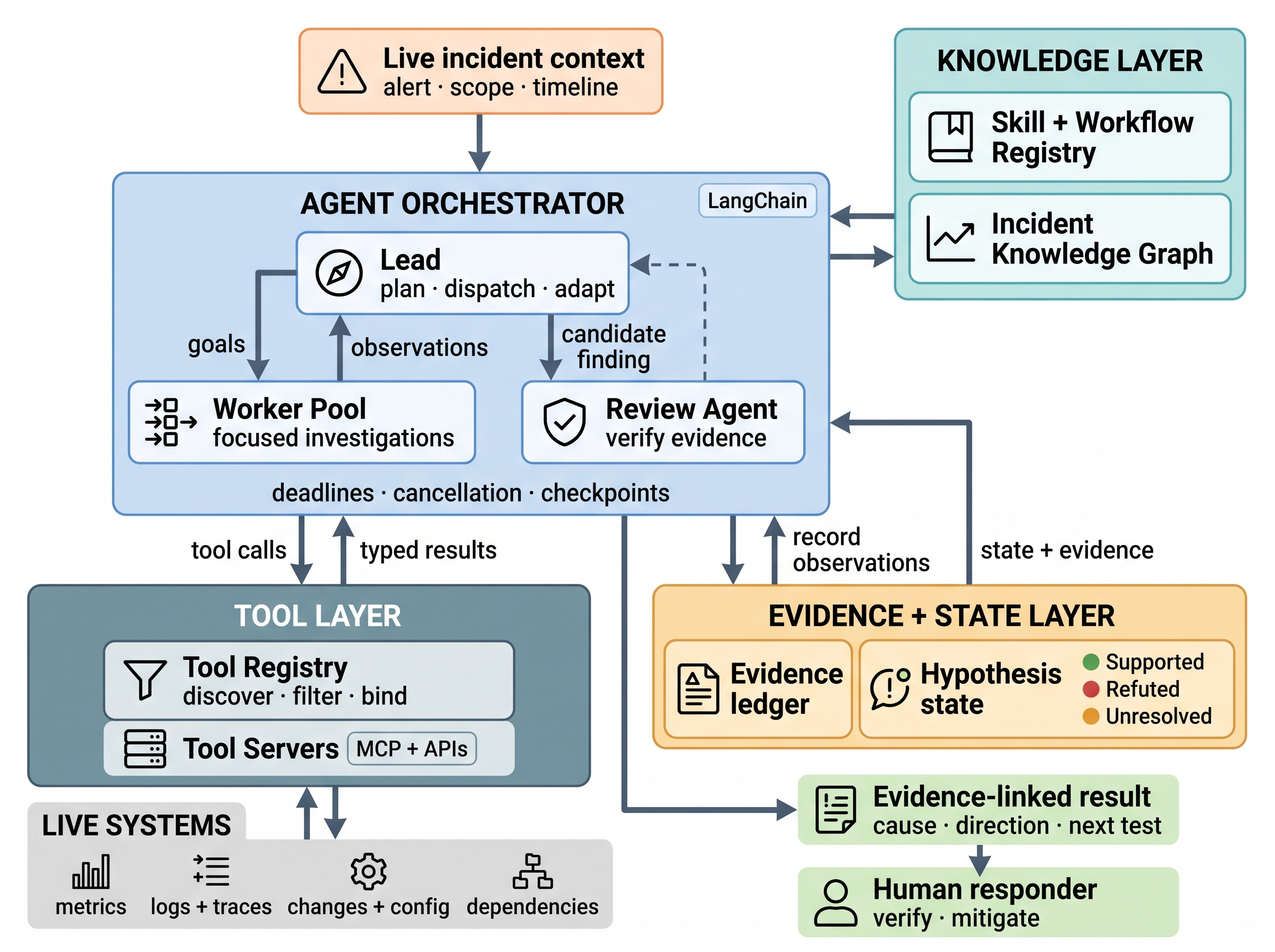}
\caption{Executing an investigation. The Lead converts incident context and
governed knowledge into focused worker goals. Workers gather live observations
through registered tools; an evidence layer preserves their source and status,
and the runtime Review Agent checks the result before it reaches a human
responder.}
\label{fig:adaptive_workflow}
\label{fig:execution_architecture}
\end{figure}

\subsection{Evaluate: Trajectories and Outcomes}
\label{sec:system_evaluation}

After mitigation, DivinerEval joins the checkpointed trajectory with a reference
outcome assembled from responder-reviewed records. A coverage gate excludes
cases without a reliable reference or reviewable trajectory.

Distinct from the runtime Review Agent, the post-mitigation Agent Judge scores
three dimensions. \emph{Exact RCA} measures whether the result identifies the
reviewed causal mechanism. \emph{Direction} measures whether it usefully narrows
the investigation toward the correct subsystem or failure mechanism, even when
the exact cause is missing. \emph{Groundedness} measures whether claims and
hypothesis changes follow cited evidence with the correct source, scope, time,
and tool outcome.

The Agent Judge emits scores, justifications, supporting or contradicting
records, and artifact-linked failure tags. Experts audit these records and
revise the versioned rubric when they find systematic errors.
\appref{app:quality_scoring} defines the complete scoring and validation
protocol.

\subsection{Adapt: Updating Operational Knowledge}
\label{sec:system_adaptation}

Adapt is a human-gated process for deriving, replaying, and publishing
operational-knowledge updates. Diviner groups recurring reviewed failures and
identifies the smallest artifact that can address them.
Missing diagnostic logic maps to a skill; missing or poorly
ordered checks to a workflow; inaccessible evidence to a tool; stale scope to a
registry entry; and missing history to the IKG. Agent Judge errors instead
update the evaluation rubric. Each proposal records its evidence, owner, scope,
and expected effect.

Candidates pass structural validation and paired replay against their baseline
using the same alert-time inputs and captured responses. Gates test the target
repair and regressions in quality, latency, and tool behavior.

Passing creates a reviewable version, not an automatic update. The owner
inspects the diff, evidence, replay results, and regressions before approving,
revising, or rejecting it. Approved versions remain monitored and reversible;
\appref{app:adaptation_protocol} gives the detailed checks.

Replay cannot reproduce live permissions, throttling, concurrency, or timing,
so later investigations remain the test of durability. The registry records
lineage from reviewed cases through the proposal to any approved artifact
version.

\section{Implementation}
\label{sec:implementation}

Diviner separates model-directed investigation from code-enforced operational
boundaries. Registries resolve applicable knowledge and tools; the runtime
coordinates the Lead, isolated workers, and runtime Review Agent; and the
evidence layer records typed observations, hypotheses, and checkpoints. Model
calls choose what to investigate and interpret returned data, while code
enforces artifact versions, tool access, evidence admission, deadlines, and
publication boundaries. The subsections follow this execution path in
Figure~\ref{fig:execution_architecture}.

\subsection{Registry-Based Governance}
\label{sec:impl_registry}
\label{sec:impl_tools}

Skills, workflows, and tools share one registry contract. Each descriptor
contains a stable identifier, owner, version, namespace, applicability scope,
provenance, enablement state, and a pointer to the full artifact. Agents search
compact descriptors first and load the selected artifact only at dispatch. This
progressive resolution avoids placing the complete catalog in every prompt and
records the exact version used by each investigation.

For example, an investigation of rejected requests may retrieve a
model-serving workflow, resolve a payload-compatibility skill, and expose only
the schema, request, and error-analysis tools required by that skill. The
dispatch records each selected version; unrelated skills and tools remain
outside the worker's context.

\parab{Selection and validation.} Workflow retrieval first filters by namespace
and applicability, then ranks the remaining descriptors against the incident
scope and preliminary symptoms. Skill retrieval uses semantic relevance with
lexical matching as a fallback and applies quality and provenance only to break
close matches. Each workflow stores expert guidance and a directed graph of
skill identifiers. Load-time validation verifies that every referenced skill
exists and is enabled. The Lead instantiates the selected workflow as the live
plan and adapts that plan as evidence arrives. These checks prevent popularity
from overriding applicability and detect stale references before an incident.

\parab{Tool views.} The registry exposes a large catalog of diagnostic tools and
workflow templates. An agent does not receive this entire surface. The
runtime creates a task-specific view from the
selected skill, a small common tool set, and any explicit step override. A
worker may search for and bind another registered tool, but the new binding and
its usage state remain local to that worker. The runtime records each tool's
invocations, successes, and failures. Offline evaluation uses these records to
update quality signals for later selection. Across artifact types, the
registries follow one pattern: progressive context loading, just-in-time
binding, per-agent resource isolation, offline quality feedback, and governance
through a shared versioned interface. \apprefs{app:skill_registry}{app:tool_plane}
give the complete registry and tool contracts.

\subsubsection{Registry Lifecycle and Audit}
\label{app:skill_registry}

Table~\ref{tab:registry_lifecycle} makes explicit which state is recorded as an
artifact moves from discovery through binding, dispatch, and update.

\begin{table}[h]
\centering
\caption{Registry lifecycle and recorded state.}
\label{tab:registry_lifecycle}
\footnotesize
\setlength{\tabcolsep}{3pt}
\begin{tabular}{@{}p{0.18\columnwidth}p{0.40\columnwidth}p{0.34\columnwidth}@{}}
\toprule
\textbf{Stage} & \textbf{Action} & \textbf{Recorded state} \\
\midrule
Discover & Search compact metadata; rank relevance first & Query and candidate identifiers \\
Bind & Resolve an enabled artifact version & Identifier and version \\
Dispatch & Construct a scoped worker context and tool view & Worker, plan step, and actual tool view \\
Update & Replay and review before publication & Proposal lineage and new version \\
\bottomrule
\end{tabular}
\end{table}

Registry membership establishes identity and review history, not correctness or
authorization. Backing systems still authorize every operational query. A
worker may bind another registered tool as evidence changes, but that binding
remains local to the worker. Disable controls remove stale artifacts from future
selection without rewriting prior trajectories.

\subsection{Agent Runtime and Adaptive Orchestration}
\label{sec:impl_models}

Diviner uses a Lead to orchestrate workers rather than delegating the
investigation lifecycle to a model-platform-native agent. The surrounding runtime
preserves consistent enforcement over tools, checkpoints, interruptions, and
evidence attribution as model platforms change. It is built with
LangChain~\cite{langchain2022}; Diviner implements the Lead orchestration,
state, and evidence contracts. Backend adapters provide every model with the
same execution context and require the same typed result format, so changing
the model does not change the plan, evidence, or evaluation schemas.
The controlled study in Section~\ref{sec:backtest_results} exercises this
boundary with multiple model backends.

\parab{Lead loop.} A tool result, worker update, or responder message reactivates
the Lead from the latest checkpoint. The Lead may add, reorder, or retire a plan
step; dispatch, steer, or cancel a worker; or propose a conclusion. Before
scheduling work, the runtime validates requested dispatches and tool calls
against the registry snapshot, permissions, and remaining budget. The Lead can
revise the investigation but cannot bypass these runtime boundaries.

\parab{Isolated dispatch.} Each worker receives one question, one skill, the
relevant incident context, and its resolved tool view. Independent workers run
concurrently. They return structured observations rather than sharing raw
transcripts, which isolates one worker's instructions, tool failures, and
temporary bindings from another worker's context. The Lead merges
their observations into the common hypothesis state.

\subsubsection{Checkpointed Execution Algorithm}
\label{app:adaptive_execution}

Diviner keeps the registered workflow, incident-specific plan, and observed
evidence separate. The workflow is a reviewed prior and cannot change during a
run. Each live-plan step states a testable question, target scope, and expected
evidence. The Lead owns and revises the plan and dispatches workers. The runtime
validates skill and tool bindings, records results, checkpoints state, and
enforces deadlines. Algorithm~\ref{alg:adaptive_execution} defines this
boundary between Lead-directed investigation and enforced runtime constraints.

\begin{algorithm}[h]
\small
\caption{Checkpointed adaptive investigation}
\label{alg:adaptive_execution}
\begin{algorithmic}[1]
\Require Incident alert and versioned registry snapshot
\State Retrieve a relevant workflow and initialize the live plan
\While{the investigation deadline has not expired}
  \State Lead proposes plan updates and registered-skill dispatches
  \State Runtime validates boundaries and runs independent workers
  \State Admit valid tool observations and checkpoint the trajectory
  \State Lead updates hypotheses, priorities, and unresolved questions
  \If{the Lead proposes a conclusion}
    \State Runtime Review Agent checks the conclusion against recorded evidence
    \If{supported}
      \State \Return an evidence-linked result for responder review
    \EndIf
    \State Add material review gaps to the live plan
  \EndIf
\EndWhile
\State \Return a bounded partial assessment with explicit gaps
\end{algorithmic}
\end{algorithm}

\subsection{Evidence, Checkpointing, and Replay}

\parab{Evidence admission.} Evidence is stored separately from generated prose.
Each record contains the observation or retrievable source, query target and
time range, backend service, worker, skill version, plan step, and typed outcome. A
finding adds an interpretation and confidence while citing one or more evidence
records. A valid empty result may count only within its recorded scope. A
timeout, denial, malformed response, or backend failure remains part of the
trace but cannot support or refute a hypothesis.

\subsubsection{Tool Invocation and Outcome Semantics}
\label{app:tool_plane}

MCP~\cite{mcp2025spec} standardizes discovery and invocation, but does not make
a tool result trustworthy. Diviner places MCP tools and native integrations
behind one contract containing the tool's identity, the backend service it
calls, its intended use, typed inputs, and policy metadata. Here, a backend
service is an operational system that supplies data or executes a query, such
as a metrics, logs, or configuration service; it is not the model provider. If
one backend service is unavailable, only the tools that depend on it are
removed from the resolved catalog, while tools backed by other services remain
available to the investigation.

\parab{Filtering and invocation.} The orchestrator begins with a curated Lead
view, adds tools declared by the selected skill, and removes mutation and
publication capabilities. Search and late binding can add another registered
tool without loading every schema initially. The backing service still enforces
identity, authorization, and data access.

For each call, the runtime resolves the selected tool's backend service,
validates the request, attaches the incident and worker attribution and a
deadline, invokes the service, and normalizes the response.
Table~\ref{tab:tool_outcomes} records the distinction that matters for
reasoning: an empty, valid query may be informative, whereas a failed query is
not negative evidence.

\begin{table}[t]
\centering
\caption{Normalized tool outcomes and their evidentiary meaning.}
\label{tab:tool_outcomes}
\footnotesize
\setlength{\tabcolsep}{3pt}
\begin{tabular}{@{}p{0.27\columnwidth}p{0.65\columnwidth}@{}}
\toprule
\textbf{Outcome} & \textbf{Interpretation} \\
\midrule
Data returned & Candidate observation after scope and provenance checks \\
Valid empty result & Potential counter-evidence only for the requested scope \\
Invalid or denied & No evidence; correct the request or report the authorization failure \\
Timeout, throttle, or backend failure & No evidence; retain the operational failure \\
\bottomrule
\end{tabular}
\end{table}

A validated read becomes an evidence record containing the observation or a
retrievable source, query scope and time range, backend service, worker, skill,
and plan step. Generated prose does not enter this path. The agent can still
choose the wrong query or misinterpret a valid result; provenance makes those
errors inspectable rather than claiming that tool use eliminates them.

\parab{Checkpointing.} After each Lead turn, Diviner checkpoints the live plan,
workers, directives, evidence, findings, attribution, and current assessment.
The event loop serializes state through an ordered asynchronous writer and
flushes it when a run pauses or ends. A run can resume on another host without
reconstructing state from generated text. The same records later provide the
trajectory consumed by DivinerEval.

\parab{Caching and replay.} Embeddings and materialized skill bundles use
bounded process-local caches. Tool-result caching is opt-in because stale
telemetry can become false evidence. Historical replay instead uses explicit
snapshots of requests and responses. Applying the baseline and candidate
artifact versions to the same captured evidence makes their behavior directly
comparable without querying a changed live system or introducing evidence that
became available only after mitigation. Replay cannot reproduce live
authorization, throttling, or cross-system timing.

\subsection{Authorization and Failure Semantics}
\label{sec:impl_reliability}

Diviner remains advisory. It does not execute mitigations, and an agent cannot
publish a proposed artifact change. The runtime removes known mutation and
publication tools from worker views. Each backing service still enforces user
identity, authorization, and data classification. This second check remains
necessary because registry membership is not an authorization grant and not
every integrated tool is globally read-only.

The runtime bounds worker lifetime, retries, concurrent dispatch, and tool
exposure. A deadline returns a partial assessment with explicit gaps rather
than forcing completion. Backend failures remain visible in the trajectory and
cannot become negative evidence. Rate limits remain enforced at backend-service
boundaries, while backend load and tail latency are monitored separately.

\section{Evaluation}
\label{sec:evals}

We evaluate Diviner through four questions:
(RQ1) Does it provide timely, evidence-grounded investigation at operational
scale?
(RQ2) Is Diviner use associated with better incident outcomes?
(RQ3) What end-to-end benefit does Diviner provide over a generic tool-using
agent, and how does performance change across agentic platforms?
and (RQ4) Can reviewed experience produce governed operational-knowledge
updates?
Each analysis uses a distinct population and unit of measurement.

For the outcome analysis, TDM plots normalize values to the final unassisted
median; relative effects and confidence intervals are computed from the
underlying unrounded observations. Table~\ref{tab:deployment_summary}
summarizes the evidence, and the following sections define the populations,
denominators, and comparison protocols.

\begin{table}[h]
\centering
\caption{Operational summary}
\label{tab:deployment_summary}
\footnotesize
\setlength{\tabcolsep}{3pt}
\begin{tabular}{@{}p{3.05cm}p{5.15cm}@{}}
\toprule
\textbf{Measure} & \textbf{Result} \\
\midrule
Observed TDM difference & Approximately 52\% lower (95\% CI: 45.7--57.9\%) \\
\midrule
Observed pre-mitigation effort & Approximately 15\% lower (95\% CI: 11.0--18.1\%) \\
\midrule
Investigation quality & Full scored operational set;
45\% Exact RCA, 60\% useful Direction, and 80\% mean Groundedness \\
\midrule
Operational agent latency & Final conclusion: approximately 15 min p50 and 24.0 min p95;
first useful, medium-confidence hypothesis: 7.0 min p50 and 13.7 min p95 \\
\midrule
Early-trigger advantage & Agent-specific alerts launch investigation before
responder-facing incident thresholds and provide roughly a half-hour head start \\
\midrule
Knowledge-update activity & $>100$ cycles; about 30 proposed new skills; all adopted changes passed replay and human approval \\
\midrule
Early post-update association & 20--40\% higher quality for updated skills;
approximately 5\% higher aggregate Direction score \\
\bottomrule
\end{tabular}
\end{table}

\subsection{Evaluation Measures and Populations}
\label{app:terminology}

Table~\ref{tab:terminology} separates end-to-end incident outcomes from agent
latency and distinguishes useful direction from exact diagnosis. Each research
question then states its eligibility rules and unit of analysis; results from
different populations are not treated as directly comparable.

\begin{table}[h]
\centering
\caption{Evaluation measures, included records, and definitions.}
\label{tab:terminology}
\footnotesize
\setlength{\tabcolsep}{2.5pt}
\begin{tabular}{@{}p{0.22\columnwidth}p{0.27\columnwidth}p{0.43\columnwidth}@{}}
\toprule
\textbf{Measure} & \textbf{Included records} & \textbf{Definition} \\
\midrule
TDM & Incidents with valid detection and mitigation times & Detection to mitigation, including investigation, coordination, and recovery \\
Final latency & Completed agent runs & End-to-end agent runtime from trigger acceptance to the final conclusion \\
First direction & Runs with fine-grained checkpoints & Trigger acceptance to the first useful, medium-confidence hypothesis \\
Exact RCA & Investigations with reviewed outcomes & Specific causal mechanism identified \\
Direction & Investigations with reviewed outcomes & Useful narrowing toward the correct subsystem or mechanism \\
Groundedness & Investigations with reviewable evidence & Degree to which claims remain supported by cited evidence \\
\bottomrule
\end{tabular}
\end{table}

\subsection{RQ1: Operational Scale, Quality, and Latency}
\label{sec:deployment_quality}

\parab{Measurement scope.}
Reach counts include recorded incident and alert invocations. Quality results
include only investigations with usable responder-reviewed outcomes, while
latency results require the corresponding execution timestamps. Cases without
sufficient reviewed evidence remain uncovered rather than being counted as
successes or failures.

\parab{Operational reach.}
Diviner has supported several thousand incidents, including several hundred
high-severity incidents, and runs several thousand alert-triggered
investigations per month across multiple teams. Trigger volume exceeds the
unique-incident count because not every alert becomes an incident and one
incident can generate multiple alerts.

\subsubsection{DivinerEval Scoring and Judge Validation}
\label{app:quality_scoring}

DivinerEval first joins the recorded trajectory with a reference outcome
assembled from responder-reviewed artifacts. The trajectory contains plan
changes, tool requests and outcomes, admitted evidence, hypotheses, findings,
and the final report. The reference contains the reviewed causal mechanism,
affected scope, contributing factors, timeline, and mitigation when available.
The reference is withheld during execution and exposed only to evaluation.
Unlike the runtime Review Agent, which verifies a candidate finding during an
active investigation, the post-mitigation Agent Judge scores a completed
trajectory against this reference.

Before scoring, a coverage gate verifies that the reference supports a causal
comparison and that the trajectory contains enough evidence to inspect the
agent's claims. Cases that fail either condition remain uncovered. The Agent
Judge then scores the dimensions in Table~\ref{tab:quality_rubric}
independently and emits a structured justification with supporting and
contradicting record identifiers.

\begin{table}[t]
\centering
\caption{DivinerEval dimensions and principal scoring tests.}
\label{tab:quality_rubric}
\footnotesize
\setlength{\tabcolsep}{3pt}
\begin{tabular}{@{}p{0.20\columnwidth}p{0.34\columnwidth}p{0.38\columnwidth}@{}}
\toprule
\textbf{Dimension} & \textbf{Positive test} & \textbf{Does not receive credit} \\
\midrule
Exact RCA & Supported conclusion matches the reviewed causal mechanism. & Restating symptoms, impact, or only the affected component. \\
Direction & Findings narrow the search toward the correct subsystem, failure family, or mechanism. & Broad guesses or elimination unsupported by current evidence. \\
Groundedness & Claims and hypothesis transitions follow cited evidence with valid source, scope, time, and outcome. & Unsupported detail or transitions, contradiction, failed-query misuse, unreliable elimination, repeated work, or inconsistent time ranges. \\
\bottomrule
\end{tabular}
\end{table}

Exact RCA, Direction, and Groundedness remain separate rather than being folded
into one success label. The Agent Judge evaluates causal agreement and cited evidence,
not lexical similarity between the report and reference. Operational Exact RCA
and Direction rates count covered cases that reach the corresponding rubric's
success band; Groundedness is reported as a mean normalized score. Controlled
comparisons report mean normalized scores within each dimension. Because
Groundedness covers both final claims and the evidence-bearing trajectory, a
correct final answer does not erase an unsupported path. To validate the Agent Judge,
two domain experts independently scored Exact RCA, Direction, and Groundedness
for a representative sample using the same rubric. They did not see the Agent Judge's
scores or agent configuration. Disagreements were resolved before the final
human ratings were compared with the Agent Judge's ratings. Systematic errors
identified through this process inform revisions to the versioned rubric.

\parab{Investigation quality.}
Across all scored investigations, Diviner identifies the exact root cause in
45\% of cases and a useful causal Direction in 60\%; its mean Groundedness score
is 80\%. Direction credits useful narrowing without presenting it as a completed
diagnosis. Groundedness is a mean rubric score over claims and trajectory steps,
not the fraction of reports containing no unsupported statement.

\parab{Where Diviner succeeds and fails.}
To expose variation hidden by aggregate scores, we map each incident's title and
reviewed cause into the taxonomy in Figure~\ref{fig:incident_categories}.
Table~\ref{tab:category_gaps} groups the resulting scores into broad performance
tiers.

\begin{table}[h]
\centering
\caption{Category-level operational performance across all available Diviner
evaluations. Scores are incident-weighted percentages. Green, amber, and red
cells denote scores of at least 60\%, 40--59\%, and below 40\%, respectively.}
\label{tab:category_gaps}
\footnotesize
\setlength{\tabcolsep}{2.5pt}
\renewcommand{\arraystretch}{1.08}
\begin{tabular}{@{}>{\raggedright\arraybackslash}p{4.55cm}
                    >{\centering\arraybackslash}p{0.72cm}
                    >{\centering\arraybackslash}p{0.72cm}
                    >{\centering\arraybackslash}p{0.85cm}@{}}
\toprule
\textbf{Root-cause category} & \textbf{RCA} & \textbf{Dir.} & \textbf{Ground.} \\
\midrule
\rowcolor{green!10}
\multicolumn{4}{@{}l}{\textbf{Stronger observed performance}} \\
Infrastructure, hardware, and network & \cellcolor{yellow!24}55\% & \cellcolor{green!18}65\% & \cellcolor{green!30}\textbf{80\%} \\
Traffic, scaling, and load management & \cellcolor{yellow!24}42\% & \cellcolor{green!18}65\% & \cellcolor{green!18}74\% \\
\addlinespace[2pt]
\rowcolor{yellow!12}
\multicolumn{4}{@{}l}{\textbf{Moderate performance}} \\
Latency and performance regression & \cellcolor{yellow!24}41\% & \cellcolor{yellow!24}55\% & \cellcolor{green!18}61\% \\
Capacity and quota deficit & \cellcolor{yellow!24}40\% & \cellcolor{yellow!24}55\% & \cellcolor{green!18}69\% \\
Code deploy and code defect & \cellcolor{yellow!24}40\% & \cellcolor{yellow!24}54\% & \cellcolor{green!18}67\% \\
\addlinespace[2pt]
\rowcolor{red!9}
\multicolumn{4}{@{}l}{\textbf{Needs improvement}} \\
Model publishing and lifecycle & \cellcolor{red!14}38\% & \cellcolor{yellow!24}46\% & \cellcolor{green!18}72\% \\
Memory leak and resource saturation & \cellcolor{red!14}30\% & \cellcolor{yellow!24}45\% & \cellcolor{green!18}66\% \\
Experiment and QE rollout & \cellcolor{red!14}25\% & \cellcolor{red!14}36\% & \cellcolor{green!18}62\% \\
\bottomrule
\end{tabular}
\end{table}

The strongest observed Direction scores occur in infrastructure and network and
in traffic-scaling investigations, where the causal chain is often visible in
direct telemetry. Capacity, latency debugging, and code changes form a middle
tier. The clearest gaps are experiment rollouts, memory leaks, and
model-lifecycle investigations.

These gaps identify missing evidence rather than only model weaknesses.
Experiment failures require time-aligned rollout and assignment histories,
followed by checks that distinguish correlation from causation. Memory and
latency investigations require resource timelines, profiles, and cross-host
comparisons. Model-lifecycle failures require publish-to-serve lineage that
connects a specific artifact to observed impact. Improving these categories
therefore begins with agent-readable datasets and typed diagnostic tools, then
adds the skills and workflow priors that interpret their results.

\parab{Investigation latency.}
We report two distinct latency endpoints. Across completed investigations in
the measurement window, Diviner reaches a
final conclusion in approximately 15 minutes at p50 and 24.0 minutes at p95.
Fine-grained checkpoints are available for a more recent subset of approximately
100 incidents. Diviner produces a first useful, medium-confidence hypothesis
after 7.0 minutes at p50 and 13.7 minutes at p95. These
measurements start when Diviner accepts the trigger and quantify agent-side
latency. Final-conclusion latency covers completed investigations, whereas the
first-direction measure uses only the more recent checkpointed subset.

\parab{Early-trigger advantage.}
To offset agent execution latency, we recently introduced agent-specific alert
triggers that launch Diviner before symptoms cross the threshold for opening a
responder-facing incident. This head start matters most for complex
investigations that require many cross-system checks. In matched incidents using
the same agent configuration, alert-triggered output arrived before
incident-triggered output in 95\% of cases, with a median advantage of roughly
half an hour; the middle half ranged from approximately a quarter hour to more
than an hour. In 40\% of cases, the alert-triggered output was available before
the incident-triggered investigation began. This result measures output
availability.

\parab{Takeaway.}
At operational scale, Diviner more often provides a useful causal direction
than an exact diagnosis, and its conclusions remain strongly grounded in
recorded evidence. It produces useful direction within minutes, while early
triggering can make that output available substantially sooner to responders.
Performance nevertheless varies by failure class and by the evidence exposed
through operational tools.

\subsection{RQ2: Impact on Incident Outcomes}
\label{sec:deployment_impact}

\begin{figure}[t]
\centering
\includegraphics[width=\columnwidth]{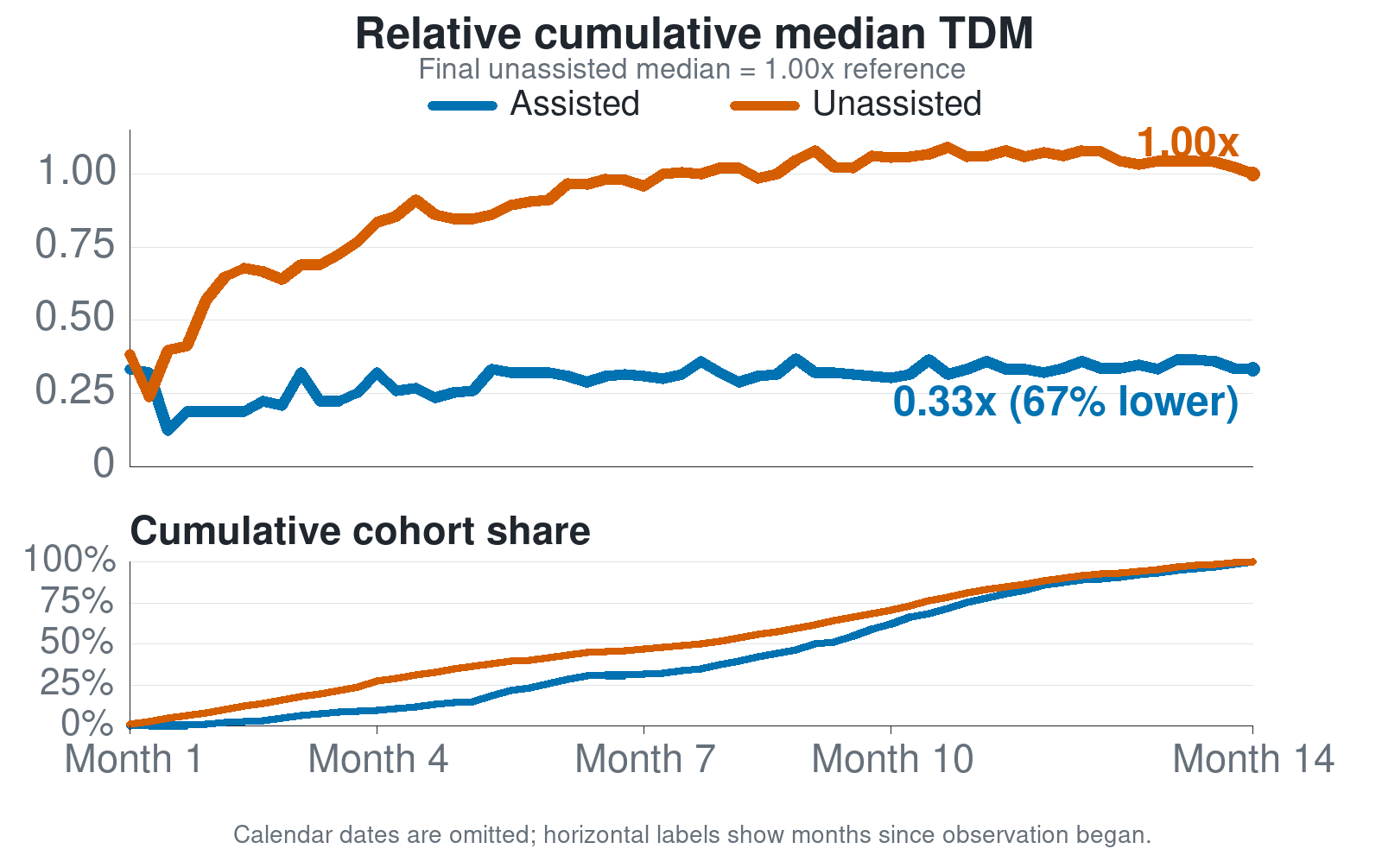}
\caption{Assisted and unassisted TDM in a fixed operational cohort. Calendar
dates are omitted. The top panel recomputes cumulative p50 TDM and normalizes
both series to the final unassisted median; the bottom panel shows cumulative
cohort share. }
\label{fig:tdm_timeline}
\end{figure}

\parab{Method.}
We select incidents with valid detection-to-mitigation time (TDM) from a fixed
operational population. An incident is classified as assisted only when its
metadata records a Diviner invocation. TDM includes investigation,
coordination, and mitigation; it is not agent runtime. Figure
\ref{fig:tdm_timeline} recomputes the cumulative median at successive
observation checkpoints. Separately defined adoption cohorts compare outcomes
before and after domain-specific skills and workflows became available.

\parab{Mitigation time.}
The fixed population contains several thousand incidents. At the latest
checkpoint, the assisted cumulative median is approximately one-third of the
unassisted median. We separately analyze adoption cohorts defined around the
availability of domain-expert-authored skills and workflows. Across onboarded
cohorts, Diviner-assisted incidents record approximately 52\% lower TDM
(95\% CI: 45.7--57.9\%). Cohorts without custom onboarding show an
approximately 0--10\% reduction. This difference is consistent with the value
of structured knowledge for RCA agents.

\parab{Operational effort.}
Diviner-assisted incidents record approximately 15\% lower pre-mitigation
operational effort (95\% CI: 11.0--18.1\%). This metric estimates engineering
effort before mitigation. The cohort comparison is observational and may also
reflect differences in adoption, incident mix, staffing, or observability.

\parab{Takeaway.}
Diviner assistance is associated with shorter mitigation time and lower
pre-mitigation engineering effort. The larger reduction in cohorts with
domain-specific skills and workflows is consistent with the value of encoded
operational knowledge.

\subsection{RQ3: End-to-End Benefit over a Generic Agent and Platform Effects}
\label{sec:backtest_results}

We first compare a generic agent with the Diviner harness while holding the
underlying model fixed. We then compare the same harness across agentic
platforms.

\subsubsection{Controlled-Study Protocol}
\label{app:backtest}

The study separates end-to-end harness benefit from platform effects. Claude
vanilla is compared with Diviner + Claude to measure the benefit of adding the
Diviner harness while holding the underlying model fixed. Diviner + Claude and
Diviner + Codex then show how the same harness performs across agentic
platforms. The operational configuration provides additional context. Every
configuration receives the same incident input and is scored against the same
responder-reviewed outcome after execution.

\begin{table}[h]
\centering
\caption{Controlled-study protocol.}
\label{tab:backtest_protocol}
\footnotesize
\setlength{\tabcolsep}{3pt}
\begin{tabular}{@{}p{0.24\columnwidth}p{0.68\columnwidth}@{}}
\toprule
\textbf{Element} & \textbf{Design choice} \\
\midrule
Baseline & Claude vanilla with native capabilities and without the Diviner harness \\
Harness comparison & Diviner + Claude adds the Diviner harness while retaining the baseline model \\
Diviner harness & Governed skills and workflows, historical retrieval, Lead--worker orchestration, tool and evidence contracts, and checkpointing \\
Platform comparison & Diviner + Codex applies the same harness through a different agentic platform \\
Operational context & An additional operational configuration is reported for context \\
Pairing & Every configuration receives the same alert-time input for each incident \\
Knowledge boundary & Incident discussion, reviewed outcomes, and prior agent analyses are withheld \\
Replay & Point-in-time tool responses are used where available \\
Measures & Exact RCA, Direction, Groundedness, and wall time \\
Analysis & Candidate-minus-baseline differences with paired-bootstrap confidence intervals \\
\bottomrule
\end{tabular}
\end{table}

The incident is the unit of pairing. For each Diviner configuration, we compute
its within-incident score and runtime differences from Claude vanilla, then
bootstrap those paired differences to form the reported 95\% intervals. The
harness-effect estimand is the difference between Diviner + Claude and Claude
vanilla. Comparisons among the Diviner configurations describe performance
variation across agentic platforms or runtime implementations. The study
evaluates the harness as a bundle and does not identify the contribution of an
individual component.

Controlled wall times use p50/p90 on this historical subset and are not directly
comparable with the operational p50/p95 latency measurements in RQ1.

\begin{figure}[t]
\centering
\includegraphics[width=\columnwidth]{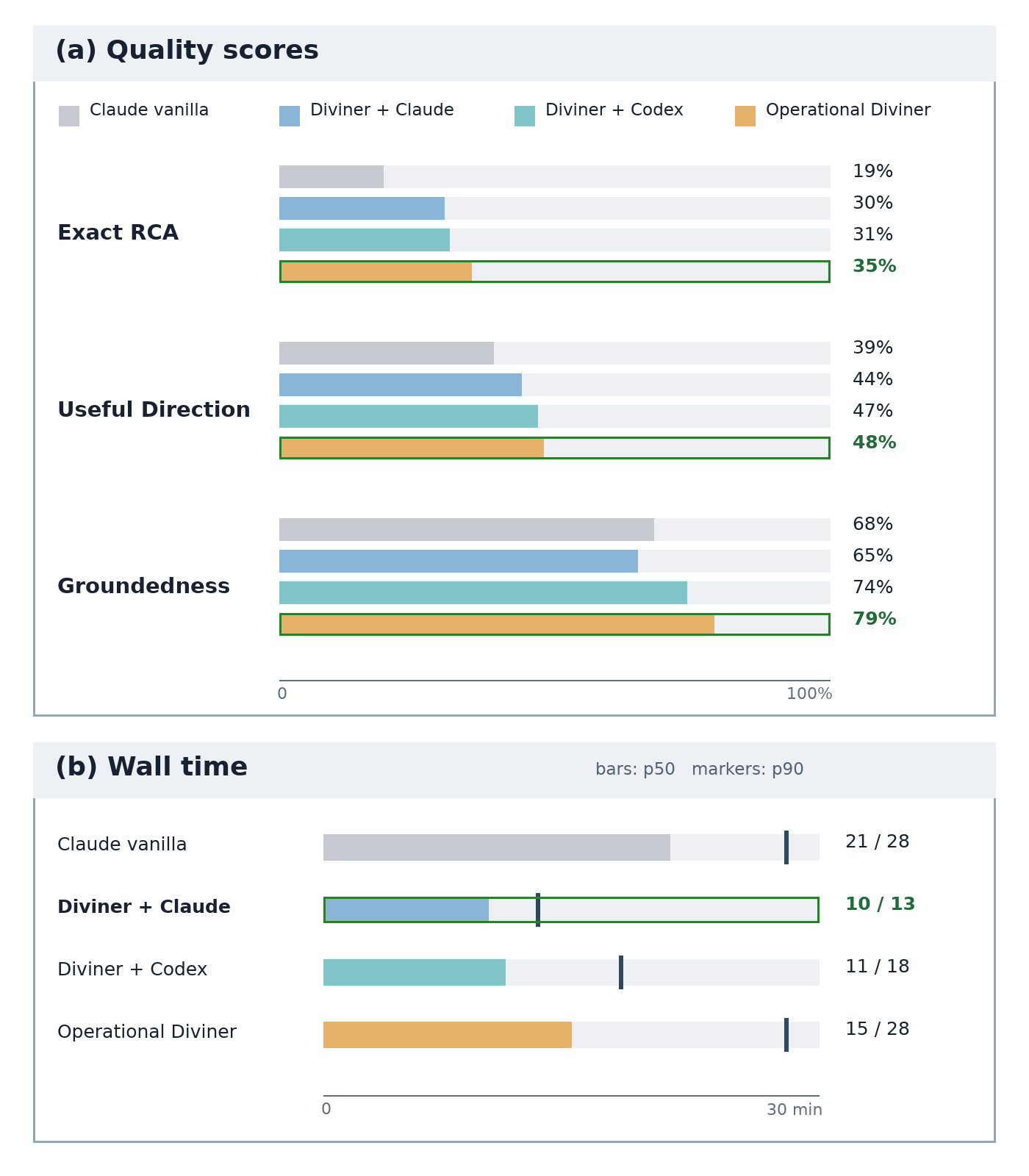}
\caption{Controlled results for the generic-agent baseline, the Diviner harness
on two agentic platforms, and the operational configuration. Panel
(a) reports mean Exact RCA, useful Direction, and Groundedness scores. Panel (b)
reports wall time, with bars showing p50 and markers showing p90. Paired
differences and confidence intervals are reported in the text.}
\label{fig:backtest_quality}
\end{figure}

\parab{End-to-end harness benefit.}
Against Claude vanilla, Diviner + Claude improves Exact RCA by 11 percentage
points (95\% CI: 3--20) and reduces mean wall time by 52\% (95\% CI: 44--58).
Direction was 5 percentage points higher and Groundedness was 3 points lower,
but neither difference was statistically distinguishable from no change.
Because the harness combines
governed knowledge, adaptive planning, evidence controls, and checkpointing,
the result does not isolate any individual mechanism.

\parab{Platform effects.}
Across the two Diviner harness configurations, Exact RCA and Direction are
similar for Claude and Codex (30\% versus 31\%, and 44\% versus 47\%), while
Codex has higher Groundedness (74\% versus 65\%) and slightly longer p50/p90 wall
time (11/18 versus 10/13 minutes). The operational configuration has point
estimates of 35\% Exact RCA, 48\% Direction, and 79\% Groundedness, with a
15/28-minute wall time; its Groundedness difference from baseline is 11
percentage points (95\% CI: 7--13). The Claude--Codex comparison describes how
the same harness performs across platforms, while the operational configuration
provides additional context. Neither comparison isolates a single model or
runtime component.\footnote{The 80\% operational Groundedness score in RQ1 and
the 79\% controlled-study point estimate use different incident populations,
evidence availability, and evaluation conditions and are not directly
comparable.}

\parab{Takeaway.}
With the underlying model held fixed, adding the Diviner harness improves Exact
RCA and runtime. The mean Direction score rises, but its interval includes no
change; Groundedness also shows no clear change. With the harness held fixed,
changing the agentic platform leaves Exact RCA and Direction broadly similar
but changes Groundedness and runtime. The study therefore establishes an
end-to-end harness benefit while exposing platform-specific tradeoffs.

\subsection{RQ4: Human-Gated Operational-Knowledge Updates}
\label{sec:adaptation_results}

Diviner implements a human-gated process for deriving, replaying, and publishing
operational-knowledge updates. The process uses reviewed corpus cases, saved
trajectories, and on-call review to identify recurring, operationally important
gaps.

\subsubsection{Update Validation Protocol}
\label{app:adaptation_protocol}

Adaptation starts from completed DivinerEval records. The pipeline groups
recurring failure tags by domain and artifact version and creates a candidate
change with explicit scope, ownership, motivating cases, and expected effect.
Table~\ref{tab:adaptation_targets} maps observed failures to candidate artifact
changes. Automated analysis proposes the target; an artifact owner may revise
or reject that classification.

\begin{table}[t]
\centering
\caption{How DivinerEval findings route to adaptation targets.}
\label{tab:adaptation_targets}
\footnotesize
\setlength{\tabcolsep}{3pt}
\begin{tabular}{@{}p{0.37\columnwidth}p{0.19\columnwidth}p{0.36\columnwidth}@{}}
\toprule
\textbf{Observed gap} & \textbf{Target} & \textbf{Candidate change} \\
\midrule
Missing or incorrect diagnostic procedure & Skill & Add checks, decision rules, or interpretation guidance \\
\midrule
Wrong ordering, dependency, or coverage & Workflow & Revise initial steps, branches, or skill bindings \\
\midrule
Evidence unavailable, malformed, or repeatedly misused & Tool contract & Add a tool or repair inputs, outputs, and failure semantics \\
\midrule
Stale scope, ownership, or eligibility & Registry & Correct metadata, filtering, enablement, or version selection \\
\midrule
Missing or misleading historical precedent & IKG & Add, relink, rescope, or retire a failure signature \\
\midrule
Systematic judge disagreement & Evaluation rubric & Refine scoring criteria without changing agent behavior \\
\bottomrule
\end{tabular}
\end{table}

Table~\ref{tab:adaptation_gates} summarizes the subsequent admission gates.

\begin{table}[t]
\centering
\caption{Checks applied before an adaptation can enter a registry.}
\label{tab:adaptation_gates}
\footnotesize
\setlength{\tabcolsep}{3pt}
\begin{tabular}{@{}p{0.22\columnwidth}p{0.70\columnwidth}@{}}
\toprule
\textbf{Gate} & \textbf{Check} \\
\midrule
Artifact validation & Schema parses; referenced artifacts resolve; owner, scope, and version are present; requested capabilities remain allowed. \\
Point-in-time replay & Baseline and candidate receive the same incident-time inputs, registry cutoff, and available tool snapshots. \\
Quality and regression & Target cases improve without material regressions in Exact RCA, Direction, Groundedness, latency, or tool behavior on relevant cases. \\
Leakage control & Runtime inputs exclude reviewed outcomes and later discussion; the candidate artifact is the only intended post-incident change. \\
Owner decision & The owner reviews the artifact diff, motivating evidence, replay comparison, and recorded regressions before approval. \\
Release control & An approved change receives a new version and retains monitoring, disablement, rollback, and proposal lineage. \\
\bottomrule
\end{tabular}
\end{table}

Replay is a comparison under preserved evidence, not a reconstruction of live
execution. It cannot reproduce authorization changes, throttling, concurrency,
or cross-system timing. A passing replay therefore makes a proposal eligible
for owner review; it never publishes the change by itself.

\parab{Observed update activity.}
The process has completed more than 100 cycles and proposed about 30 new skills.
Among proposals that passed validation and replay and reached an artifact owner,
90\% were accepted. This rate measures final-stage approval, not acceptance of
raw model-generated proposals. Among skills that completed an update and
re-evaluation cycle, measured quality was 20--40\% higher after the update.
Across operational evaluations, aggregate Direction rose by approximately 5\%.

Only replayed, owner-approved versions become available to later
investigations.

These are early before-and-after measurements from a limited operating period;
the number of post-change incidents remains small, and incident mix and
evaluation coverage can change between versions. The results establish that the
pipeline can derive, evaluate, and publish reviewed artifact updates. They do
not establish that those updates improve future investigations. That stronger
claim requires a temporally held-out evaluation comparing old and updated
artifact versions on incidents excluded from proposal generation and replay.

\parab{Takeaway.}
Diviner can convert reviewed failures into replay-tested, owner-approved
knowledge updates. These measurements demonstrate a governed update process,
not autonomous self-improvement or durable causal gains.

\section{Operational Lessons}
\label{sec:lessons}

This section summarizes four lessons from operating Diviner: how incident
records should be prepared for reuse, why investigation latency matters, what
historical replay must preserve, and how investigation quality should be
evaluated.

\parab{Incident reports are not ready-to-use memory.}
Postmortems combine observations available during response with diagnoses and
actions recorded later. Giving an agent the full report can reveal the answer,
add irrelevant details, and bias it toward a previous cause. Diviner instead
converts each reviewed incident into a compact failure signature containing its
scope, symptoms, relevant change, causal mechanism, distinguishing evidence,
and mitigation. Automated processing links the signature to normalized service
and failure-family entities. During a new incident, retrieved cases suggest
possible mechanisms and diagnostic checks, but current evidence must still
confirm them. This turns a recurring pattern into a focused starting point
without treating a past conclusion as evidence for the present incident.

\parab{A correct answer can arrive too late.}
For familiar failures, an experienced responder may follow a known diagnostic
path faster than an agent. The agent must select tools, transfer data, retry
failed calls, and complete several model turns. A correct diagnosis delivered
after mitigation provides little value during the response, so latency is part
of investigation quality. Diviner starts from earlier agent-specific alerts,
runs independent checks concurrently, and uses typed tools that can answer
several related queries in one call. It also saves checkpoints so responders
can inspect partial progress. Because early alerts are more frequent and can be
noisy, execution budgets and backend limits control the resulting load.

\parab{Replay must preserve what was known at the time.}
Replaying only the incident description is not enough. Telemetry may expire,
and later changes to permissions, schemas, configurations, or topology can
alter both the available evidence and its meaning. A valid replay therefore
uses the original alert input, saved tool responses, and the skill, workflow,
knowledge-graph, model, registry, configuration, and judge versions available
at that time. Postmortems and reviewed root-cause fields are withheld because
they contain conclusions reached later. Even with these controls, replay cannot
recover evidence that was never saved or reproduce live authorization,
throttling, latency, and consistency. We therefore use it to compare system
versions, not to recreate an incident exactly.

\parab{Quality includes the investigation path.}
Final-answer accuracy misses useful progress, such as ruling out a subsystem or
identifying the correct causal direction. It can also reward a lucky answer
that relies on stale, failed, or unrelated evidence. Prompt injection or leaked
conclusions can similarly produce a correct-looking answer that the saved
trajectory does not support. Diviner therefore scores
Exact RCA, Direction, and Groundedness separately over the saved trajectory.
These measures help distinguish failures in hypothesis selection, evidence
collection, and synthesis, so reviewers can update the relevant skill,
workflow, tool, or evaluation rule. Domain experts audit the Agent Judge and
revise its rubric when needed.

\section{Related Work}
\label{sec:related}

\parab{Coding-agent harnesses and operational RCA.}
General agent frameworks provide reasoning-and-action loops, tool binding, and
graph-based orchestration~\cite{yao2023react,langchain2022,langgraph1210}.
Unlike coding tasks with repeatable build or test signals, operational RCA
begins from a partial symptom in a changing, permissioned environment without a
deterministic verifier. On-call evaluation likewise finds that coding ability
does not directly transfer~\cite{gong2026oncall}. Diviner adds governed
operational knowledge, typed evidence, point-in-time replay, and trajectory
evaluation.

\parab{Telemetry-based RCA.}
DeepLog models log sequences; Eadro and DiagFusion fuse telemetry; ESRO
retrieves causes from outage reports; and KRCA combines causal-graph drilldown
with multi-agent verification
\cite{du2017deeplog,lee2023eadro,zhang2023diagfusion,chakraborty2023esro,jiang2026krca}.
These systems primarily localize faults within a defined topology; Diviner
investigates across models, configurations, shared infrastructure, and history.

\parab{LLM-based incident agents.}
RCACopilot combines incident context with operational handlers; Nissist executes
troubleshooting guides; RCAgent performs tool-augmented diagnosis; and Xpert
recommends diagnostic queries
\cite{chen2024rcacopilot,an2024nissist,wang2024rcagent,jiang2024xpert}.
AgentRCA maintains competing hypotheses and gathers statistical evidence
\cite{wei2026agentrca}. Diviner additionally governs which local knowledge and
observations are permitted, how trajectories are evaluated, and which reviewed
experience may change later behavior.

\parab{Knowledge and adaptation.}
OpsHarness evolves operational knowledge through trajectory mining and staged
verification, while SkillForge generates and repairs domain-grounded skills
\cite{huang2026opsharness,liu2026skillforge}. RAG, GraphRAG, and Voyager provide
retrieval, structured memory, and reusable skills
\cite{lewis2020rag,edge2024graphrag,wang2023voyager}. Diviner separates
procedures, precedent, live evidence, and runtime state by authority and update
rate, turning reviewed experience into versioned proposals rather than silent
prompt or model changes.

\parab{Agent control and interoperability.}
Diviner is built with LangChain but uses its own Lead-directed orchestration
loop as evidence changes. Agent Skills and MCP standardize procedures and tools
\cite{agentskills2026spec,mcp2025spec}; Diviner combines them with scoped
registries, authorization, typed outcomes, deadlines, evidence schemas, and
checkpoints.

\parab{Evaluation.}
OpenRCA, AIOpsLab, and ITBench provide reproducible RCA and IT-agent environments
\cite{xu2025openrca,chen2025aiopslab,itbench2025}; a recent on-call benchmark
adds live telemetry and expert-calibrated judging~\cite{gong2026oncall}.
MT-Bench and Agent-as-a-Judge study model evaluation, while Reflexion uses feedback
\cite{zheng2023judging,zhuge2024agentasajudge,shinn2023reflexion}. Diviner
adds sustained operational evidence and scores exact cause, useful direction,
and Groundedness over the complete trajectory.

\section{Conclusion}
\label{sec:conclusion}

Diviner is an agentic RCA system for large-scale distributed infrastructure.
It combines operational knowledge derived from reviewed incidents and
service-owner expertise with an adaptive Lead--worker runtime. During an
incident, Diviner starts from an applicable workflow, adjusts its investigation
as evidence arrives, and returns an evidence-linked cause, useful direction, or
next test. Afterward, reviewed gaps can produce replay-tested knowledge updates,
but artifact owners decide whether those updates become available to future
investigations.

Across scored operational investigations, Diviner identifies the exact root
cause in 45\% of cases, provides a useful causal direction in 60\%, and achieves
80\% mean Groundedness. In a controlled same-model comparison, adding the
Diviner harness improves Exact RCA and reduces runtime. Mean Direction rises,
but its interval includes no change; Groundedness shows no clear change.
Observational cohort comparisons associate Diviner assistance with shorter
mitigation time and lower pre-mitigation engineering effort, but do not isolate
a causal effect.

The central lesson is that reliable RCA agents require more than model
capability and tool access. They need governed operational knowledge,
point-in-time evidence, bounded execution, evaluation of the investigation
path, and human control over durable changes. Diviner's experience shows how
these elements can work together to provide useful incident assistance and to
update operational knowledge through reviewed experience.

\bibliographystyle{ACM-Reference-Format}
\bibliography{references}

\end{document}